\documentclass[runningheads]{llncs}
\usepackage[T1]{fontenc}
\usepackage{graphicx}
\usepackage{booktabs} % for midrule
\usepackage{multirow} % for multirow
\usepackage{amsmath}
\usepackage{amsfonts}
\usepackage{float}
\begin{document}
%
% \title{Task-Specific Distillation for Plant Species \\and Disease Recognition}
\title{Energy-Efficient Plant Monitoring \\via Knowledge Distillation}

% \author{Anonymous Author(s)}
% \authorrunning{Anonymous et al.}
% \institute{Anonymous Institution(s)}
%%%%%%%%%%%%
\author{Ilyass Moummad\inst{1}\orcidID{0009-0003-2925-2500} \and
Reda Bensaid\inst{2}\orcidID{0009-0008-5369-0857} \and
Kawtar Zaher\inst{1,3}\orcidID{0009-0008-9920-9483} \and
Hervé Goëau\inst{4}\orcidID{0000-0003-3296-3795} \and
Jean-Christophe Lombardo\inst{1}\orcidID{0000-0002-9656-4219} \and
Joseph Salmon\inst{5}\orcidID{0000-0002-3181-0634} \and
Pierre Bonnet\inst{4}\orcidID{0000-0002-2828-4389} \and
Alexis Joly\inst{1}\orcidID{0000-0002-2161-9940}}
\authorrunning{I. Moummad et al.}
\institute{LIRMM, Univ Montpellier, CNRS, Inria, Montpellier, France \\
% \url{http://www.springer.com/gp/computer-science/lncs} 
\and
IMT Atlantique, Brest, France. Polytechnique Montréal, Canada\\
\and
Institut National de l’Audiovisuel, Paris, France
\and
AMAP, Université de Montpellier, CIRAD, INRAE, IRD, Montpellier, France\\
\and
IMAG, Univ Montpellier, CNRS, Inria, Montpellier, France\\
\email{\{ilyass.moummad,alexis.joly\}@inria.fr}}
\maketitle              % typeset the header of the contribution

\begin{abstract}
Recent advances in large-scale visual representation learning have significantly improved performance in plant species and plant disease recognition tasks. However, state-of-the-art models, often based on high-capacity vision transformers or multimodal foundation models, remain computationally expensive and difficult to deploy in resource-constrained environments such as mobile or edge devices. This limitation hinders the scalability of automated biodiversity monitoring and precision agriculture systems, where efficiency is as critical as accuracy.

In this work, we investigate knowledge distillation as an effective approach to transfer the representational capacity of large pretrained models into smaller, more efficient architectures. We focus on plant species and disease recognition, and conduct an extensive empirical study on two challenging benchmarks: Pl@ntNet300K-v2 and Deep-Plant-Disease. We evaluate four representative architectures, including two ConvNeXt models and two vision transformers, under multiple training regimes: from-scratch training and pretrained initialization, each with and without distillation. In total, we train and evaluate 70 models.

Our results show that knowledge distillation consistently improves performance across tasks and architectures. Distilled models are able to match the performance of significantly larger models while maintaining substantially lower computational cost. These findings demonstrate the potential of knowledge distillation techniques to enable efficient and scalable deployment of plant recognition systems in real-world environmental applications.\footnote{Code and models: \url{https://github.com/ilyassmoummad/distillplant}}

\end{abstract}

\section{Introduction}

The rapid decline of biodiversity and the increasing prevalence of plant diseases pose major challenges for ecosystems, agriculture, and global food security~\cite{disease}. Scalable monitoring solutions are therefore essential to support biodiversity conservation, ecological research, and precision agriculture~\cite{monitoring}. In this context, automated visual recognition systems have emerged as a key enabling technology, allowing non-experts and large-scale platforms to identify plant species and diagnose diseases directly from images~\cite{plantnetapp}.

Recent progress in computer vision has significantly advanced the capabilities of such systems. In particular, large-scale pretrained models, often based on transformer architectures, have demonstrated strong performance in fine-grained visual recognition tasks~\cite{bioclip2}. By leveraging massive datasets and high-capacity architectures, these models are able to capture subtle visual differences between species and disease patterns~\cite{inquire}. As a result, modern systems such as large vision encoders~\cite{dinov2} and deployed applications like Pl@ntNet~\cite{plantnetapp} achieve high accuracy across diverse environmental conditions.

However, these gains come at a substantial computational cost~\cite{greenai}. High-capacity models require significant memory and processing resources, making them difficult to deploy in practical scenarios such as mobile devices, embedded platforms, or in-field sensors. This limitation is especially critical for environmental monitoring, where scalability, energy efficiency, and accessibility are key constraints~\cite{wildlifeperspectives}. Consequently, there is a growing need for approaches that retain the accuracy of large models while enabling efficient inference.

Model compression techniques provide a natural direction to address this challenge~\cite{efficientdl}. Among them, knowledge distillation has emerged as a flexible framework for transferring information from a large teacher model to a smaller student model~\cite{kd}. Rather than training compact models solely from labeled data, distillation leverages the richer supervisory signal provided by the teacher, enabling improved generalization and performance. While this paradigm has been successfully applied in several computer vision domains, its application in fine-grained plant recognition and disease classification settings remains unexplored.

In this work, we investigate the impact of knowledge distillation on plant species and disease recognition, using benchmarks such as Pl@ntNet300K-v2~\cite{pn300k} and Deep-Plant-Disease~\cite{dpd}, which encompass large-scale biodiversity data and disease-specific visual patterns. We evaluate different architectures including convolutional networks~\cite{convnext} and vision transformers~\cite{vit} and various training regimes, combining pretrained initialization with distillation.

Our study provides an exhaustive empirical evaluation, comparing standard training with distillation under both from-scratch and pretrained settings. We aim to better understand how knowledge distillation influences modern visual representations and architectural choices in plant monitoring tasks, and to assess whether compact models can retain the discriminative capabilities of larger models. Our work seeks to demonstrate that distillation can help develop efficient models that offer competitive performance while reducing computational requirements, supporting the deployment of sustainable computer vision systems for biodiversity monitoring and plant health assessment.

\section{Related Work}

\paragraph{Automatic plant monitoring.}
The development of automated plant monitoring systems has accelerated in recent years~\cite{plantnetapp}, driven by the need for scalable biodiversity assessment and accessible tools for species identification. Applications such as ObsIdentify~\cite{obsidentify}, Flora Incognita~\cite{floraincognita}, iNaturalist~\cite{inaturalist} and Pl@ntNet~\cite{plantnetapp} have demonstrated the potential of large-scale visual recognition systems to support both citizen science and expert-driven ecological studies. These systems typically rely on high-capacity deep learning models trained on massive datasets, enabling robust performance across diverse environmental conditions and species distributions~\cite{pn300k}. More recently, advances in large pretrained visual encoders and multimodal models have further improved recognition accuracy, particularly in fine-grained settings such as plant species identification and disease diagnosis~\cite{plantclef2024,bioclip2}. However, these approaches often depend on computationally intensive architectures, limiting their deployment in real-world scenarios where low-latency and energy-efficient inference are required. In contrast to prior work that prioritizes accuracy through increasingly large models, our work focuses on improving the efficiency of plant recognition systems by transferring the capabilities of such models into smaller architectures via knowledge distillation~\cite{kd}, with the goal of enabling practical deployment in resource-constrained environments.

\paragraph{Plant recognition datasets.}
The progress of plant recognition methods has been closely tied to the availability of large and diverse annotated datasets. General-purpose biodiversity datasets such as iNaturalist~\cite{inquire} provide large-scale collections of species observations in natural conditions, supporting the development of robust fine-grained classifiers. More specialized datasets have been introduced to address domain-specific challenges: Pl@ntNet300K-v2~\cite{pn300k}\footnote{\url{https://zenodo.org/records/10419064}} focuses on plant species recognition from user-contributed images with high variability in viewpoint and quality; Deep-Plant-Disease~\cite{dpd}\footnote{\url{https://zenodo.org/records/16879271}} targets disease classification with controlled and field conditions; PlantDoc~\cite{pd} and PlantWild~\cite{pw} introduce challenging real-world disease and species recognition scenarios with complex backgrounds; and the Pl@ntCLEF benchmark series provides standardized evaluation protocols for plant identification at scale~\cite{plantclef2024}. %While these datasets have enabled significant advances, most prior work evaluates models primarily in terms of accuracy, often overlooking computational constraints. In this work, we leverage representative datasets from both species recognition and disease classification domains to systematically evaluate the trade-off between performance and efficiency, positioning our study as a bridge between high-accuracy benchmarks and deployable solutions.

\paragraph{Model compression and knowledge distillation.}
Model compression techniques aim to reduce the computational and memory footprint of deep neural networks while preserving performance~\cite{efficientdl}. Among the most widely studied approaches are pruning~\cite{pruning}, which removes redundant parameters; quantization~\cite{quantization}, which reduces numerical precision for faster inference; and knowledge distillation~\cite{kd}, which transfers information from a large teacher model to a smaller student model. Knowledge distillation, in particular, has shown strong empirical success in both convolutional and transformer architectures, benefiting from richer supervisory signals such as soft targets or intermediate feature alignment~\cite{unic,marrie}. Recent works have explored distillation in the context of self-supervised learning and large pretrained models, further enhancing its effectiveness~\cite{dinov2}. However, the application of these techniques to fine-grained plant recognition tasks remains relatively underexplored, especially in scenarios combining modern architectures and diverse training regimes. In this work, we focus on knowledge distillation as a flexible and architecture-agnostic approach to model compression, and provide an extensive empirical evaluation of its effectiveness across multiple models and training settings for plant species and disease recognition.

\section{Method}

\begin{figure}[h]
  \centering
  \includegraphics[width=1.\columnwidth]{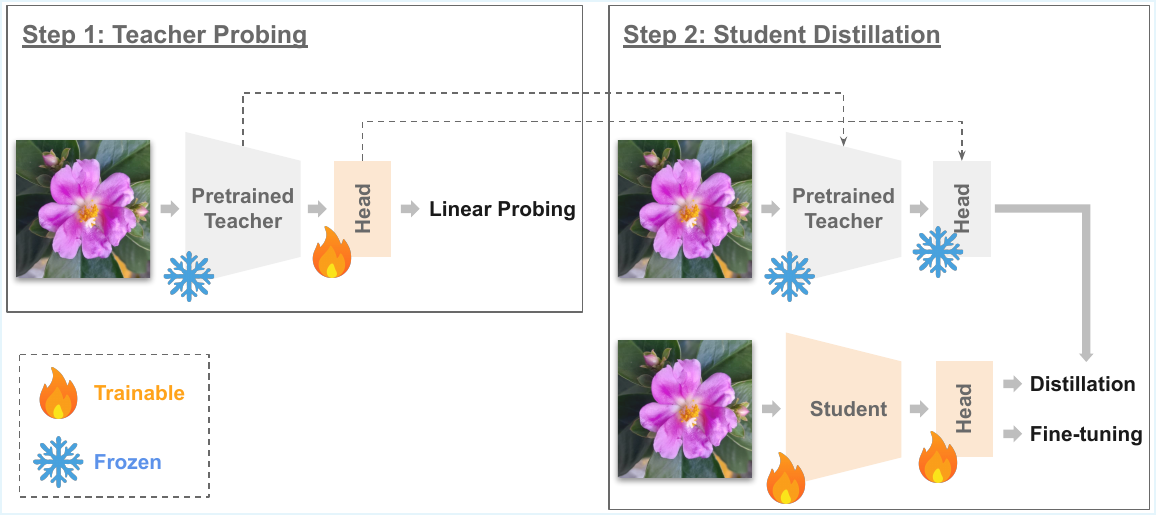}
  \caption{Two-step knowledge distillation pipeline. \textbf{Step 1:} A pretrained teacher model is adapted to the downstream task via linear probing. \textbf{Step 2:} A student model is trained to solve the task while aligning with the teacher's predictions.}
  \label{fig:distill}
\end{figure}

We aim to train efficient models for plant species and disease recognition by transferring knowledge from large pretrained visual models. To this end, we adopt a task-specific knowledge distillation framework inspired by Marrie et al.~\cite{marrie}, in which a high-capacity teacher model is first adapted to the downstream task, and a compact student model is subsequently trained to reproduce its predictions. The overall pipeline is illustrated in Figure~\ref{fig:distill}.

\subsection{Problem formulation}

Let $\mathcal{D} = \{(x_1, y_1),\dots,(x_n,y_n)\}$ be a labeled dataset of plant images, where $x_i$'s are images and $y_i$'s their class label (species or disease). A model $f$ is typically trained by minimizing a supervised objective:
\begin{equation}
\mathcal{L}_{\text{task}}(f) = \mathbb{E}_{(x,y)\sim \mathcal{D}} \ \ell_{\text{task}}(f(x), y),
\end{equation}
where $\ell_{\text{task}}$ is the cross-entropy loss for multi-class classification problems. Rather than training a compact model directly with this objective, we introduce a teacher–student framework to guide learning using a stronger model.

\subsection{Teacher adaptation}

We start from a pretrained encoder $e_t$ and construct a task-specific teacher model by attaching a linear prediction head $p_t$:
\begin{equation}
f_t(x) = p_t(e_t(x)).
\end{equation}

The teacher is adapted to the downstream task by training only a linear classification head (fully connected layer followed by softmax) while keeping the encoder frozen. This linear probing strategy is computationally efficient and preserves the general representations learned during pretraining, which have been shown to be effective for distillation~\cite{marrie}.

\subsection{Student learning via distillation}

Given the adapted teacher $f_t$, we train a compact student model $f_s$ to both predict ground-truth labels and mimic the teacher's outputs. The training objective combines both the supervised task and the distillation task:
\begin{equation}
\mathcal{L}(f_s) = (1 - \alpha)\mathcal{L}_{\text{task}}(f_s) + \alpha \mathcal{L}_{\text{distill}}(f_s, f_t),
\end{equation}
where $\alpha$ balances the contribution of each term.

The distillation loss encourages the student to match the softened output distribution of the teacher following~\cite{kd}:
\begin{equation}
\mathcal{L}_{\text{distill}}(f_s, f_t) = \mathbb{E}_{x \sim \mathcal{D}} \ T^2 \, D_{\text{KL}}\big( \sigma(f_s(x)/T) \,\|\, \sigma(f_t(x)/T) \big),
\end{equation}
where $T$ is a temperature parameter, $\sigma$ is the softmax function, and $D_{\text{KL}}$ denotes the Kullback–Leibler divergence. This formulation is architecture-agnostic and enables effective knowledge transfer across heterogeneous model families.

\subsection{Training settings}

To evaluate the impact of distillation across realistic scenarios, we consider multiple training configurations. Specifically, we vary (i) the student initialization (training from scratch or from pretrained weights), (ii) the training objective (standard supervised learning or distillation), (iii) the teacher model (supervised, self-supervised, or multimodal), and (iv) the student architecture, including both convolutional networks and vision transformers.

This experimental design allows us to systematically evaluate how distillation interacts with modern pretrained visual representations and architectural choices in plant species and disease recognition tasks.

\section{Experiments}

% downstream datasets : Pl@ntNet300K / Deep-Plant-Disease
% teacher models : Pl@ntCLEF 2024, BioCLIP 2, DINOv3 ViT-Large
% teacher probing 30 epochs (following Marrie et al.) linear lr 1e-4 min lr 1e-6 cosine decay adamw wd 1e-4
% student models : CNX-S CNX-T / ViT-S ViT-S+
% training from scratch vs dinov3 initialization, classification task, distillation task for the 3 models.
% implem details similar to teacher hyperparams, we train 70 models, hyperparam search is costly and waste of energy, we choose this set of hyperparams that is stable and where 30 epochs is enough to converge.

We evaluate the effectiveness of knowledge distillation for the recognition of plant species and diseases in multiple architectures, datasets, and training regimes. Our experiments are designed to assess: (i) whether distillation improves performance over standard fine-tuning, (ii) how it interacts with pretrained initialization, and (iii) whether compact models can approach the performance of large teacher models.

\subsection{Experimental Setup}

\paragraph{Datasets.}
We conduct experiments on two large-scale and complementary benchmarks. Pl@ntNet300K-v2~\cite{pn300k} is a fine-grained plant species recognition dataset composed of images collected in real-world conditions, exhibiting high intra-class variability and long-tailed distributions. Deep-Plant-Disease~\cite{dpd} focuses on plant disease classification and includes diverse disease categories across multiple plant species. Together, these datasets capture both biodiversity monitoring and agricultural use cases.

\begin{table}[h]
\centering
\caption{Summary of evaluation datasets.}
% \resizebox{\columnwidth}{!}{
\begin{tabular}{l c c c c}
\hline
\textbf{Dataset} & \textbf{\#Train} & \textbf{\#Val} & \textbf{\#Test} & \textbf{\#Classes} \\
\hline
Pl@ntNet300K-v2 & 243{,}866 & 31{,}115 & 31{,}106 & 1{,}000 \\
Deep-Plant-Disease & 198{,}710 & -- & 49{,}866 & 175 \\
\hline
\end{tabular}
% }
\label{tab:datasets}
\end{table}

\paragraph{Teacher models.}
We consider three high-capacity pretrained models as teachers, selected for their strong performance on plant-related visual tasks. BioCLIP-2~\cite{bioclip2} is a ViT-L model trained with a contrastive language-image pretraining objective on a large-scale dataset structured around biological taxonomy, providing rich multimodal representations. Pl@ntCLEF 2024~\cite{plantclef2024} is a ViT-B model pretrained specifically for plant species classification on the flora of southwestern Europe, making it suitable for fine-grained plant recognition. Finally, DINOv3~\cite{dinov3} ViT-L is a self-supervised model pretrained on large-scale image data, producing high-quality visual representations that capture general-purpose features without relying on labels, that are potentially useful for plant recognition tasks. All teacher models are adapted to downstream tasks using a linear probing strategy, where a linear classification head is trained on top of frozen features. This approach yields strong yet computationally efficient teacher models for task-specific knowledge distillation.

\paragraph{Student models.}
We evaluate four compact architectures representative of modern vision models: ConvNeXt-Tiny (CNX-T), ConvNeXt-Small (CNX-S), ViT-Small (ViT-S), and ViT-S+. For each architecture, we consider two initialization strategies. First, models are trained from scratch, with weights initialized from a truncated normal distribution with standard deviation 0.02. Second, models are initialized from DINOv3 pretrained weights obtained via self-supervised learning on the large-scale LVD-1689M dataset\footnote{\url{https://huggingface.co/collections/facebook/dinov3}}. All student models are trained under two regimes: standard supervised fine-tuning, and knowledge distillation combined with supervised fine-tuning.

\paragraph{Computational complexity.}
Table~\ref{tab:complexity} summarizes the number of parameters and estimated inference cost measured in GFLOPs (Giga Floating-Point Operations per second) for both student and teacher models at a $224\times224$ resolution. The compact student models require substantially fewer resources than the larger teacher models, underscoring their practical efficiency for deployment in resource-constrained scenarios.

\begin{table}[h]
\centering
\caption{Model complexity at $224\times224$ resolution.}
\begin{tabular}{l c c}
\hline
\textbf{Model} & \textbf{\#Params} & \textbf{GFLOPs} \\
\hline
\multicolumn{3}{c}{\textit{Student models}} \\
\hline
ConvNeXt-T & 29M & 3.8 \\
ConvNeXt-S & 50M & 8.4 \\
ViT-S & 21M & 9.2 \\
ViT-S+ & 29M & 12.3 \\
\hline
\multicolumn{3}{c}{\textit{Teacher models}} \\
\hline
ViT-B & 86M & 36.0 \\
ViT-L & 300M & 125.0 \\
\hline
\end{tabular}
\label{tab:complexity}
\end{table}

\paragraph{Training details.}
Both teacher and student models are trained for 30 epochs using the AdamW optimizer with a learning rate of $10^{-4}$, a minimum learning rate of $10^{-6}$ following a cosine decay schedule, and a weight decay of $10^{-4}$. Teachers are trained via linear probing on frozen features, while students are trained end-to-end. To ensure a fair comparison across 70 model configurations, we adopt this single stable training setup without hyperparameter tuning, which is sufficient to achieve reliable convergence within 30 epochs. For distillation, we follow Marrie et al.~\cite{marrie}, using a temperature $T = 2$ and a weighting coefficient $\alpha = 0.5$. 

\subsection{Results on Pl@ntNet300K-v2}

Table~\ref{tab:pn_results} reports Top-1 (micro) accuracy on Pl@ntNet300K-v2. Teacher models achieve comparable performance around $86\%$, providing strong supervision signals.

\begin{table*}[htbp]
\centering
\small
\setlength{\tabcolsep}{4pt}
\begin{tabular}{l l l l c}
\toprule
\textbf{Model} & \textbf{Init} & \textbf{Strategy} & \textbf{Teacher} & \textbf{Top-1 (\%)} \\
\midrule

\multicolumn{5}{c}{\textbf{Teachers}} \\
\midrule
BioCLIP-2 (300M)  & TreeOfLife-200M & Linear Probe & -- & 86.8 \\
Pl@ntCLEF (86M)  & Pl@ntCLEF2024   & Linear Probe & -- & 86.1 \\
DINOv3-L (300M)   & LVD-1689M       & Linear Probe & -- & 86.1 \\

\midrule
\multicolumn{5}{c}{\textbf{Students}} \\
\midrule

\multirow{8}{*}{ConvNeXt-T (29M)}
 & Scratch     & Finetune & --         & 54.8 \\
 & Scratch     & Distill  & BioCLIP-2  & 58.2 \\
 & Scratch     & Distill  & PlantCLEF  & 56.9 \\
 & Scratch     & Distill  & DINOv3-L   & 57.7 \\
 \cmidrule{2-5}
 & LVD-1689M   & Finetune & --         & 81.9 \\
 & LVD-1689M   & Distill  & BioCLIP-2  & 85.8 \\
 & LVD-1689M   & Distill  & PlantCLEF  & 85.4 \\
 & LVD-1689M   & Distill  & DINOv3-L   & 85.5 \\

\midrule

\multirow{8}{*}{ConvNeXt-S (50M)}
 & Scratch     & Finetune & --         & 60.8 \\
 & Scratch     & Distill  & BioCLIP-2  & 63.2 \\
 & Scratch     & Distill  & PlantCLEF  & 63.7 \\
 & Scratch     & Distill  & DINOv3-L   & 62.9 \\
 \cmidrule{2-5}
 & LVD-1689M   & Finetune & --         & 83.1 \\
 & LVD-1689M   & Distill  & BioCLIP-2  & 86.3 \\
 & LVD-1689M   & Distill  & PlantCLEF  & 85.9 \\
 & LVD-1689M   & Distill  & DINOv3-L   & 85.9 \\

\midrule

\multirow{8}{*}{ViT-S (21M)}
 & Scratch     & Finetune & --         & 68.3 \\
 & Scratch     & Distill  & BioCLIP-2  & 71.3 \\
 & Scratch     & Distill  & PlantCLEF  & 70.6 \\
 & Scratch     & Distill  & DINOv3-L   & 70.1 \\
 \cmidrule{2-5}
 & LVD-1689M   & Finetune & --         & 83.1 \\
 & LVD-1689M   & Distill  & BioCLIP-2  & 85.5 \\
 & LVD-1689M   & Distill  & PlantCLEF  & 85.2 \\
 & LVD-1689M   & Distill  & DINOv3-L   & 85.3 \\

\midrule

\multirow{8}{*}{ViT-S+ (29M)}
 & Scratch     & Finetune & --         & 66.9 \\
 & Scratch     & Distill  & BioCLIP-2  & 71.8 \\
 & Scratch     & Distill  & PlantCLEF  & 72.3 \\
 & Scratch     & Distill  & DINOv3-L   & 72.4 \\
\cmidrule{2-5}
 & LVD-1689M   & Finetune & --         & 82.9 \\
 & LVD-1689M   & Distill  & BioCLIP-2  & 85.9 \\
 & LVD-1689M   & Distill  & PlantCLEF  & 85.8 \\
 & LVD-1689M   & Distill  & DINOv3-L   & 85.7 \\

\bottomrule
\end{tabular}
\caption{Task-specific distillation of large pretrained visual models on Pl@ntNet300K-v2. Teacher models are trained via linear probing. Student models are trained either with fine-tuning or distillation from teacher logits, using either random initialization or DINOv3 LVD-1689M initialization.}
\label{tab:pn_results}
\end{table*}

\paragraph{Effect of distillation.}
Knowledge distillation consistently improves performance over standard fine-tuning across all architectures and initialization settings. Gains are particularly pronounced when training from scratch. For instance, ConvNeXt-T improves from $54.8\%$ to $58.2\%$ when distilled from BioCLIP-2, while ViT-S+ improves from $66.9\%$ to $72.4\%$ when distilled from DINOv3-L. These results indicate that distillation provides a strong learning signal in low-data or low-prior settings.

\paragraph{Interaction with pretrained initialization.}
When initialized from DINOv3, all models benefit from a strong performance boost, reaching above $80\%$ accuracy even without distillation. However, distillation still provides consistent improvements, typically in the range of $+2$ to $+4$ points. Notably, ConvNeXt-S improves from $83.1\%$ to $86.3\%$ when distilled from BioCLIP-2, effectively matching teacher-level performance.

% \paragraph{Comparison across teachers.}
% While all teachers provide useful supervision, BioCLIP 2 tends to yield slightly stronger results across architectures, suggesting that domain-specific multimodal pretraining may produce more informative soft targets for plant recognition tasks.

\subsection{Feature space analysis}

\begin{figure}[h]
  \centering
  \includegraphics[width=1.\columnwidth]{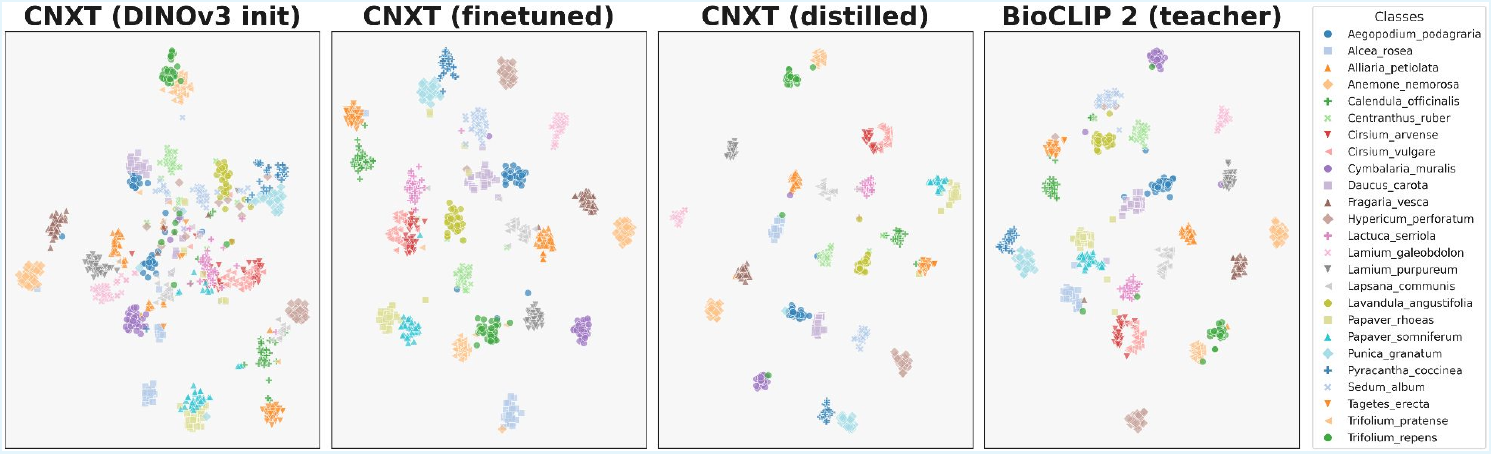}
  \caption{t-SNE visualization of feature embeddings on the PlantNet MetaAlbum dataset (Pl@ntNet300K subset with 25 most populous classes) for different models. The CNX-T model initialized from DINOv3 produces relatively dispersed and overlapping clusters, indicating limited separability of plant species. Fine-tuning improves cluster compactness and class separation, while distillation further enhances the structure of the embedding space, yielding tighter and more discriminative clusters. Interestingly, the distilled model achieves better class separation than the teacher, BioCLIP 2, suggesting that task-specific distillation can improve upon the representations of large pretrained models for fine-grained plant recognition.}
  \label{fig:tsne}
\end{figure}

For qualitative analysis, we use a subset of the Pl@ntNet300K dataset from the MetaAlbum benchmark~\cite{metaalbum}, restricted to the 25 most populous classes following BioCLIP~\cite{bioclip2}, to visualize feature representations. We project the learned embeddings using t-SNE on this subset (Figure~\ref{fig:tsne}). CNX-T initialized from pretrained DINOv3 features exhibits limited class separability, with overlapping clusters across species. Fine-tuning improves the structure of the embedding space, leading to more compact and better-separated clusters, while distillation further enhances this effect by producing tighter clusters and clearer class boundaries. Interestingly, the distilled student exhibits more structured representations than the teacher in this setting, suggesting that task-specific distillation can refine feature organization beyond what is directly inherited from large pretrained models.

\subsection{Results on Deep-Plant-Disease}

Table~\ref{tab:dpd_results} presents results on Deep-Plant-Disease. Compared to Pl@ntNet300K-v2 (Table~\ref{tab:pn_results}), overall accuracy is lower for teacher models, reflecting the distribution shift and increased complexity of disease classification relative to the pretraining data.

\begin{table*}[htbp]
\centering
\small
\setlength{\tabcolsep}{4pt}
\begin{tabular}{l l l l c}
\toprule
\textbf{Model} & \textbf{Init} & \textbf{Strategy} & \textbf{Teacher} & \textbf{Top-1 (\%)} \\
\midrule

\multicolumn{5}{c}{\textbf{Teachers}} \\
\midrule
BioCLIP-2 (300M)  & TreeOfLife-200M & Linear Probe & -- & 74.2 \\
Pl@ntCLEF (86M)  & Pl@ntCLEF2024   & Linear Probe & -- & 72.9 \\
DINOv3-L (300M)   & LVD-1689M       & Linear Probe & -- & 76.6 \\

\midrule
\multicolumn{5}{c}{\textbf{Students}} \\
\midrule

\multirow{8}{*}{ConvNeXt-T (29M)}
 & Scratch     & Finetune & --         & 66.5 \\
 & Scratch     & Distill  & BioCLIP-2  & 68.0 \\
 & Scratch     & Distill  & PlantCLEF  & 66.0 \\
 & Scratch     & Distill  & DINOv3-L   & 67.3 \\
 \cmidrule{2-5}
 & LVD-1689M   & Finetune & --         & 79.9 \\
 & LVD-1689M   & Distill  & BioCLIP-2  & 82.7 \\
 & LVD-1689M   & Distill  & PlantCLEF  & 80.7 \\
 & LVD-1689M   & Distill  & DINOv3-L   & 82.4 \\

\midrule

\multirow{8}{*}{ConvNeXt-S (50M)}
 & Scratch     & Finetune & --         & 67.7 \\
 & Scratch     & Distill  & BioCLIP-2  & 69.1 \\
 & Scratch     & Distill  & PlantCLEF  & 68.4 \\
 & Scratch     & Distill  & DINOv3-L   & 70.1 \\
 \cmidrule{2-5}
 & LVD-1689M   & Finetune & --         & 80.4 \\
 & LVD-1689M   & Distill  & BioCLIP-2  & 83.0 \\
 & LVD-1689M   & Distill  & PlantCLEF  & 81.0 \\
 & LVD-1689M   & Distill  & DINOv3-L   & 82.8 \\

\midrule

\multirow{8}{*}{ViT-S (21M)}
 & Scratch     & Finetune & --         & 73.0 \\
 & Scratch     & Distill  & BioCLIP-2  & 75.6 \\
 & Scratch     & Distill  & PlantCLEF  & 74.5 \\
 & Scratch     & Distill  & DINOv3-L   & 74.8 \\
 \cmidrule{2-5}
 & LVD-1689M   & Finetune & --         & 80.5 \\
 & LVD-1689M   & Distill  & BioCLIP-2  & 82.2 \\
 & LVD-1689M   & Distill  & PlantCLEF  & 80.7 \\
 & LVD-1689M   & Distill  & DINOv3-L   & 82.1 \\

\midrule

\multirow{8}{*}{ViT-S+ (29M)}
 & Scratch     & Finetune & --         & 73.6 \\
 & Scratch     & Distill  & BioCLIP-2  & 76.1 \\
 & Scratch     & Distill  & PlantCLEF  & 75.2 \\
 & Scratch     & Distill  & DINOv3-L   & 75.7 \\
 \cmidrule{2-5}
 & LVD-1689M   & Finetune & --         & 80.9 \\
 & LVD-1689M   & Distill  & BioCLIP-2  & 82.4 \\
 & LVD-1689M   & Distill  & PlantCLEF  & 80.8 \\
 & LVD-1689M   & Distill  & DINOv3-L   & 82.3 \\

\bottomrule
\end{tabular}
\caption{Task-specific distillation of large pretrained visual models on Deep-Plant-Disease, following the same experimental setup as in Table~\ref{tab:pn_results}.}
\label{tab:dpd_results}
\end{table*}

\paragraph{Effect of distillation.}
Distillation again leads to consistent improvements across all architectures. For example, ViT-S improves from $73.0\%$ to $75.6\%$ when distilled from BioCLIP-2, while ConvNeXt-S improves from $67.7\%$ to $70.1\%$ when distilled from DINOv3-L. These gains confirm that distillation is effective beyond species recognition and generalizes to disease classification.

\paragraph{Pretraining vs distillation.}
Pretrained initialization remains a strong factor, with all models exceeding $80\%$ accuracy when initialized from DINOv3. However, distillation continues to provide additional gains, albeit smaller than in the scratch setting. This suggests that distillation complements pretrained representations rather than replacing them.

\paragraph{Teacher comparison.}
In contrast to Pl@ntNet300K-v2, DINOv3-L emerges as the strongest teacher on this dataset, yielding the highest accuracy among probed models. While teacher probing alone underperforms compared to student fine-tuning (e.g., DINOv3-L linear probe achieves $76.6\%$ versus $80.5\%$ for a ViT-S student initialized from LVD-1689M), it still provides valuable guidance. Distillation from these teachers consistently improves student performance, indicating that even suboptimal teacher probes can benefit student models when aligned with the target domain.

\subsection{Discussion}
Across both datasets, our experiments show that knowledge distillation is a robust and effective strategy for improving compact models on plant-related tasks. It consistently enhances performance across architectures, training regimes, and domains. While pretrained initialization provides the largest single boost in accuracy, distillation systematically complements it, enabling smaller models to match or surpass the performance of much larger systems. These results highlight the practical value of distillation for deploying efficient and accurate models in real-world environmental monitoring scenarios, particularly when large pretrained models are available as teachers and are costly to fine-tune.

\section{Conclusion}

In this work, we explored knowledge distillation as a practical approach to reduce the computational footprint of modern plant recognition systems while preserving high predictive performance. Large pretrained models have become central to biodiversity monitoring and plant disease recognition, but their deployment at scale remains constrained by their cost in terms of memory, computation, and energy. Our results show that distillation provides an effective mechanism to transfer their capabilities into significantly smaller models, enabling more efficient inference without sacrificing much accuracy.

Through a comprehensive empirical study across datasets, architectures, and training regimes, we demonstrated that distilled models consistently outperform their non-distilled counterparts and can match the performance of much larger teacher models. Importantly, these gains hold across both species recognition and disease classification tasks, and complement strong pretrained initializations. This suggests that distillation is not only a performance-enhancing technique, but also a key tool for improving the efficiency–accuracy trade-off in real-world applications. From a sustainability perspective, these findings highlight the role of knowledge distillation in enabling scalable environmental monitoring systems. By reducing model size and inference cost, distillation facilitates deployment on resource-constrained platforms such as mobile devices, embedded sensors, or edge computing infrastructures, which are essential for large-scale and continuous biodiversity observation.

Looking forward, combining distillation with other model compression techniques, such as pruning and quantization, represents a promising direction to further decrease energy consumption and hardware requirements. Such integrated approaches could lead to highly optimized models tailored for specific deployment constraints, enabling energy-efficient plant recognition systems. Overall, this work contributes to the development of more sustainable machine learning approaches for biodiversity monitoring, and we hope it will encourage further research at the intersection of model efficiency, computer vision, and environmental applications.

\subsubsection{Acknowledgements} This project was funded by the French National Research Agency (ANR) through the grant Pl@ntAgroEco 22-PEAE-0009.

% \subsubsection{Acknowledgements} This work was supported by the Pl@ntAgroEco project, funded by the "Agence Nationale de la Recherche" (ANR) under the France 2030 program, within the “Agroécologie et Numérique” initiative (reference ANR-22-PEAE-0009). The authors gratefully acknowledge this support.


\bibliographystyle{splncs04}
\bibliography{mybibliography}

@article{disease,
  title={The global burden of pathogens and pests on major food crops},
  author={Savary, Serge and Willocquet, Laetitia and Pethybridge, Sarah Jane and Esker, Paul and McRoberts, Neil and Nelson, Andy},
  journal={Nature ecology \& evolution},
  volume={3},
  number={3},
  pages={430--439},
  year={2019},
  publisher={Nature Publishing Group UK London}
}

@article{monitoring,
  title={Monitoring plant functional diversity from space},
  author={Jetz, Walter and Cavender-Bares, Jeannine and Pavlick, Ryan and Schimel, David and Davis, Frank W and Asner, Gregory P and Guralnick, Robert and Kattge, Jens and Latimer, Andrew M and Moorcroft, Paul and others},
  journal={Nature plants},
  volume={2},
  number={3},
  pages={16024},
  year={2016},
  publisher={Nature Publishing Group}
}

@inproceedings{plantnetapp,
  title={{Pl@ ntnet mobile app}},
  author={Go{\"e}au, Herv{\'e} and Bonnet, Pierre and Joly, Alexis and Baki{\'c}, Vera and Barbe, Julien and Yahiaoui, Itheri and Selmi, Souheil and Carr{\'e}, Jennifer and Barth{\'e}l{\'e}my, Daniel and Boujemaa, Nozha and others},
  booktitle={Proceedings of the 21st ACM international conference on Multimedia},
  pages={423--424},
  year={2013}
}

@inproceedings{bioclip2,
    title={Bio{CLIP} 2: Emergent Properties from Scaling Hierarchical Contrastive Learning},
    author={Jianyang Gu and Samuel Stevens and Elizabeth G Campolongo and Matthew J Thompson and Net Zhang and Jiaman Wu and Andrei Kopanev and Zheda Mai and Alexander E. White and James Balhoff and Wasila Dahdul and Daniel Rubenstein and Hilmar Lapp and Tanya Berger-Wolf and Wei-Lun Chao and Yu Su},
    booktitle={The Thirty-ninth Annual Conference on Neural Information Processing Systems},
    year={2025},
}

@article{dinov2,
  title={{DINOv2: Learning Robust Visual Features without Supervision}},
  author={Oquab, Maxime and Darcet, Timoth{\'e}e and Moutakanni, Th{\'e}o and Vo, Huy and Szafraniec, Marc and Khalidov, Vasil and Fernandez, Pierre and Haziza, Daniel and Massa, Francisco and El-Nouby, Alaaeldin and Assran, Mahmoud and Ballas, Nicolas and Galuba, Wojciech and Howes, Russell and Huang, Po-Yao and Li, Shang-Wen and Misra, Ishan and Rabbat, Michael and Sharma, Vasu and Synnaeve, Gabriel and Xu, Hu and J{\'e}gou, Herv{\'e} and Bojanowski, Piotr and LeCun, Yann and Caron, Mathilde},
  journal={Transactions on Machine Learning Research (TMLR)},
  year={2023}
}

@article{inquire,
  title={{INQUIRE: A natural world text-to-image retrieval benchmark}},
  author={Vendrow, Edward and Pantazis, Omiros and Shepard, Alexander and Brostow, Gabriel and Jones, Kate E and Mac Aodha, Oisin and Beery, Sara and Van Horn, Grant},
  journal={Advances in Neural Information Processing Systems},
  volume={37},
  pages={126500--126514},
  year={2024}
}

@article{greenai,
  title={{Green AI}},
  author={Schwartz, Roy and Dodge, Jesse and Smith, Noah A and Etzioni, Oren},
  journal={Communications of the ACM},
  volume={63},
  number={12},
  pages={54--63},
  year={2020},
  publisher={ACM New York, NY, USA}
}

@article{wildlifeperspectives,
  title={Perspectives in machine learning for wildlife conservation},
  author={Tuia, Devis and Kellenberger, Benjamin and Beery, Sara and Costelloe, Blair R and Zuffi, Silvia and Risse, Benjamin and Mathis, Alexander and Mathis, Mackenzie W and Van Langevelde, Frank and Burghardt, Tilo and others},
  journal={Nature communications},
  volume={13},
  number={1},
  pages={792},
  year={2022},
  publisher={Nature Publishing Group UK London}
}

@article{efficientdl,
  title={{Efficient Deep Learning: A Survey on Making Deep Learning Models Smaller, Faster, and Better}},
  author={Menghani, Gaurav},
  journal={ACM Computing Surveys},
  volume={55},
  number={12},
  pages={1--37},
  year={2023},
  publisher={ACM New York, NY}
}

@article{kd,
  title={{Distilling the Knowledge in a Neural Network}},
  author={Hinton, Geoffrey and Vinyals, Oriol and Dean, Jeff},
  journal={arXiv preprint arXiv:1503.02531},
  year={2015}
}

@inproceedings{pn300k,
  author    = {Camille Garcin and Alexis Joly and Pierre Bonnet and Antoine Affouard and Jean-Christophe Lombardo and Mathias Chouet and Maximilien Servajean and Titouan Lorieul and Joseph Salmon},
  title     = {{Pl@ntNet-300K}: a plant image dataset with high label ambiguity and a long‑tailed distribution},
  booktitle = {Proceedings of the Neural Information Processing Systems Track on Datasets and Benchmarks},
  year      = {2021},
}

@inproceedings{dpd,
    author = {Chai, Abel Yu Hao and Jee, Kelly Li Zhen and Lee, Sue Han and Tay, Fei Siang and Vandeputte, Jules and Goeau, Herv\'{e} and Bonnet, Pierre and Joly, Alexis},
    title = {Deep-Plant-Disease Dataset Is All You Need for Plant Disease Identification},
    year = {2025},
    isbn = {9798400720352},
    publisher = {Association for Computing Machinery},
    address = {New York, NY, USA},
    booktitle = {Proceedings of the 33rd ACM International Conference on Multimedia},
    pages = {12578–12584},
    numpages = {7},
    location = {Dublin, Ireland},
    series = {MM '25}
}

@inproceedings{convnext,
  title={{A ConvNet for the 2020s}},
  author={Liu, Zhuang and Mao, Hanzi and Wu, Chao-Yuan and Feichtenhofer, Christoph and Darrell, Trevor and Xie, Saining},
  booktitle={Proceedings of the IEEE/CVF conference on computer vision and pattern recognition},
  pages={11976--11986},
  year={2022}
}

@inproceedings{vit,
    title={{An Image is Worth 16x16 Words: Transformers for Image Recognition at Scale}},
    author={Alexey Dosovitskiy and Lucas Beyer and Alexander Kolesnikov and Dirk Weissenborn and Xiaohua Zhai and Thomas Unterthiner and Mostafa Dehghani and Matthias Minderer and Georg Heigold and Sylvain Gelly and Jakob Uszkoreit and Neil Houlsby},
    booktitle={International Conference on Learning Representations},
    year={2021},
    url={https://openreview.net/forum?id=YicbFdNTTy}
}

@misc{obsidentify,
  author       = {{Observation International}},
  title        = {ObsIdentify: Wildlife and plant identification app},
  year         = {2026},
  howpublished = {\url{https://observation.org/apps/obsidentify/}},
}

@article{floraincognita,
  title        = {{The Flora Incognita app – Interactive plant species identification}},
  author       = {M{\"a}der, Patrick and Boho, David and Rzanny, Michael and Seeland, Marco and Wittich, Hans Christian and Deggelmann, Alice and W{\"a}ldchen, Jana},
  journal      = {Methods in Ecology and Evolution},
  year         = {2021},
  volume       = {12},
  number       = {7},
  pages        = {1335--1342},
}

@misc{inaturalist,
  author       = {{iNaturalist community}},
  title        = {iNaturalist},
  howpublished = {\url{https://www.inaturalist.org}},
  year         = {2025},
}

@inproceedings{plantclef2024,
  title={{Overview of PlantCLEF 2024: Multi-species plant identification in vegetation plot images}},
  author={Goeau, Hervé and Espitalier, Vincent and Bonnet, Pierre and Joly, Alexis},
  booktitle={Working Notes of CLEF 2024 -- Conference and Labs of the Evaluation Forum},
  series={CEUR Workshop Proceedings},
  year={2024}
}

@incollection{pd,
  title={{PlantDoc: A Dataset for Visual Plant Disease Detection}},
  author={Singh, Davinder and Jain, Naman and Jain, Pranjali and Kayal, Pratik and Kumawat, Sudhakar and Batra, Nipun},
  booktitle={Proceedings of the 7th ACM IKDD CoDS and 25th COMAD},
  pages={249--253},
  year={2020}
}

@inproceedings{pw,
  title={{Benchmarking In-the-Wild Multimodal Plant Disease Recognition and A Versatile Baseline}},
  author={Wei, Tianqi and Chen, Zhi and Huang, Zi and Yu, Xin},
  booktitle={Proceedings of the 32nd ACM International Conference on Multimedia},
  pages={1593--1601},
  year={2024}
}

@article{pruning,
  title={{Optimal Brain Damage}},
  author={LeCun, Yann and Denker, John and Solla, Sara},
  journal={Advances in neural information processing systems},
  volume={2},
  year={1989}
}

@article{quantization,
  title={{Compressing Deep Convolutional Networks using Vector Quantization}},
  author={Gong, Yunchao and Liu, Liu and Yang, Ming and Bourdev, Lubomir},
  journal={arXiv preprint arXiv:1412.6115},
  year={2014}
}

@article{marrie,
    title={{On Good Practices for Task-Specific Distillation of Large Pretrained Visual Models}},
    author={Juliette Marrie and Michael Arbel and Julien Mairal and Diane Larlus},
    journal={Transactions on Machine Learning Research},
    issn={2835-8856},
    year={2024},
}

@inproceedings{unic,
    title={{{UNIC}: Universal Classification Models via Multi-teacher Distillation}},
    author={Sariyildiz, Mert Bulent and Weinzaepfel, Philippe and Lucas, Thomas and Larlus, Diane and Kalantidis, Yannis},
    booktitle={European Conference on Computer Vision (ECCV)},
    year={2024},
}

@article{metaalbum,
  title={{Meta-Album: Multi-domain Meta-Dataset forFew-Shot Image Classification}},
  author={Ullah, Ihsan and Carri{\'o}n-Ojeda, Dustin and Escalera, Sergio and Guyon, Isabelle and Huisman, Mike and Mohr, Felix and Van Rijn, Jan N and Sun, Haozhe and Vanschoren, Joaquin and Vu, Phan Anh},
  journal={Advances in Neural Information Processing Systems},
  volume={35},
  pages={3232--3247},
  year={2022}
}

@misc{dinov3,
  title={{DINOv3}},
  author={Sim{\'e}oni, Oriane and Vo, Huy V. and Seitzer, Maximilian and Baldassarre, Federico and Oquab, Maxime and Jose, Cijo and Khalidov, Vasil and Szafraniec, Marc and Yi, Seungeun and Ramamonjisoa, Micha{\"e}l and Massa, Francisco and Haziza, Daniel and Wehrstedt, Luca and Wang, Jianyuan and Darcet, Timoth{\'e}e and Moutakanni, Th{\'e}o and Sentana, Leonel and Roberts, Claire and Vedaldi, Andrea and Tolan, Jamie and Brandt, John and Couprie, Camille and Mairal, Julien and J{\'e}gou, Herv{\'e} and Labatut, Patrick and Bojanowski, Piotr},
  year={2025},
  eprint={2508.10104},
  archivePrefix={arXiv},
  primaryClass={cs.CV},
  url={https://arxiv.org/abs/2508.10104},
}

\end{document}